\pdfoutput=1

\documentclass[11pt]{article}
\usepackage[]{acl}
\usepackage{hyperref}
\usepackage{nameref}
\usepackage{times}
\usepackage{latexsym}

\usepackage[T1]{fontenc}

\usepackage[utf8]{inputenc}

\usepackage{microtype}

\usepackage{inconsolata}

\usepackage{CJKutf8}

\usepackage{enumitem}

\usepackage{graphicx}
\usepackage{pgfplots}
\pgfplotsset{width=\textwidth,compat=1.9}

\usepackage{longtable}
\usepackage{tabu}
\usepackage{multirow}
\usepackage{multicol}
\usepackage{float}
\usepackage{cuted}
\usepackage{makecell}
\usepackage{booktabs}
\usepackage{tabularx}

\usepackage{amsmath}
\usepackage{amsthm}
\usepackage{amssymb}
\usepackage{amsfonts}
\usepackage{cancel}
\usepackage{bbding}

\usepackage{algorithm,algorithmic}

\usepackage{url}

\definecolor{red}{RGB}{254,76,97}
\definecolor{orange}{RGB}{243,156,17}
\definecolor{yellow}{RGB}{255,193,22}
\definecolor{green}{RGB}{82,196,26}
\definecolor{cyan}{RGB}{52,152,219}
\definecolor{blue}{RGB}{40,120,181}
\definecolor{purple}{RGB}{157,61,207}
\definecolor{black}{RGB}{14,29,105}
\definecolor{gray}{RGB}{191,191,191}
\definecolor{DDDD}{RGB}{0,0,0}

\title{Evaluation is All You Need: Strategic Overclaiming of LLM Reasoning Capabilities Through Evaluation Design}

\author{
\vspace{-2em}
\\
Yongfu Zhu\thanks{$^{*}$ Equal contribution.},
Lin Sun\footnotemark[1],
Jinzhu Wu\footnotemark[1],
Weihong Lin\footnotemark[1], 
Xiaoqi Jian, 
Guangxiang Zhao, 
\\ 
Change Jia, 
Linglin Zhang, 
Sai-er Hu, 
Yuhan Wu,
Xiangzheng Zhang
}

\begin{document}
    \maketitle
    \begin{strip}
        \begin{center}
            \begin{minipage}{0.9\textwidth}
                \vspace{-7em}
\begin{abstract}
    \vspace{1em}
    \large
    Reasoning models represented by the Deepseek-R1-Distill series have been widely adopted by the open-source community due to their strong performance in mathematics, science, programming, and other domains. However, our study reveals that their benchmark evaluation results are subject to significant fluctuations caused by various factors as shown in Figure~\ref{fig: figure1}. Subtle differences in evaluation conditions can lead to substantial variations in results. Similar phenomena are observed in other open-source inference models fine-tuned based on the Deepseek-R1-Distill series, as well as in the QwQ-32B model, making their claimed performance improvements difficult to reproduce reliably. Therefore, we advocate for the establishment of a more rigorous paradigm for model performance evaluation and present our empirical assessments of the Deepseek-R1-Distill series models.
\end{abstract}

\vspace{2em}
\begin{figure}[H]
\centering
    \includegraphics[width=0.96\linewidth]{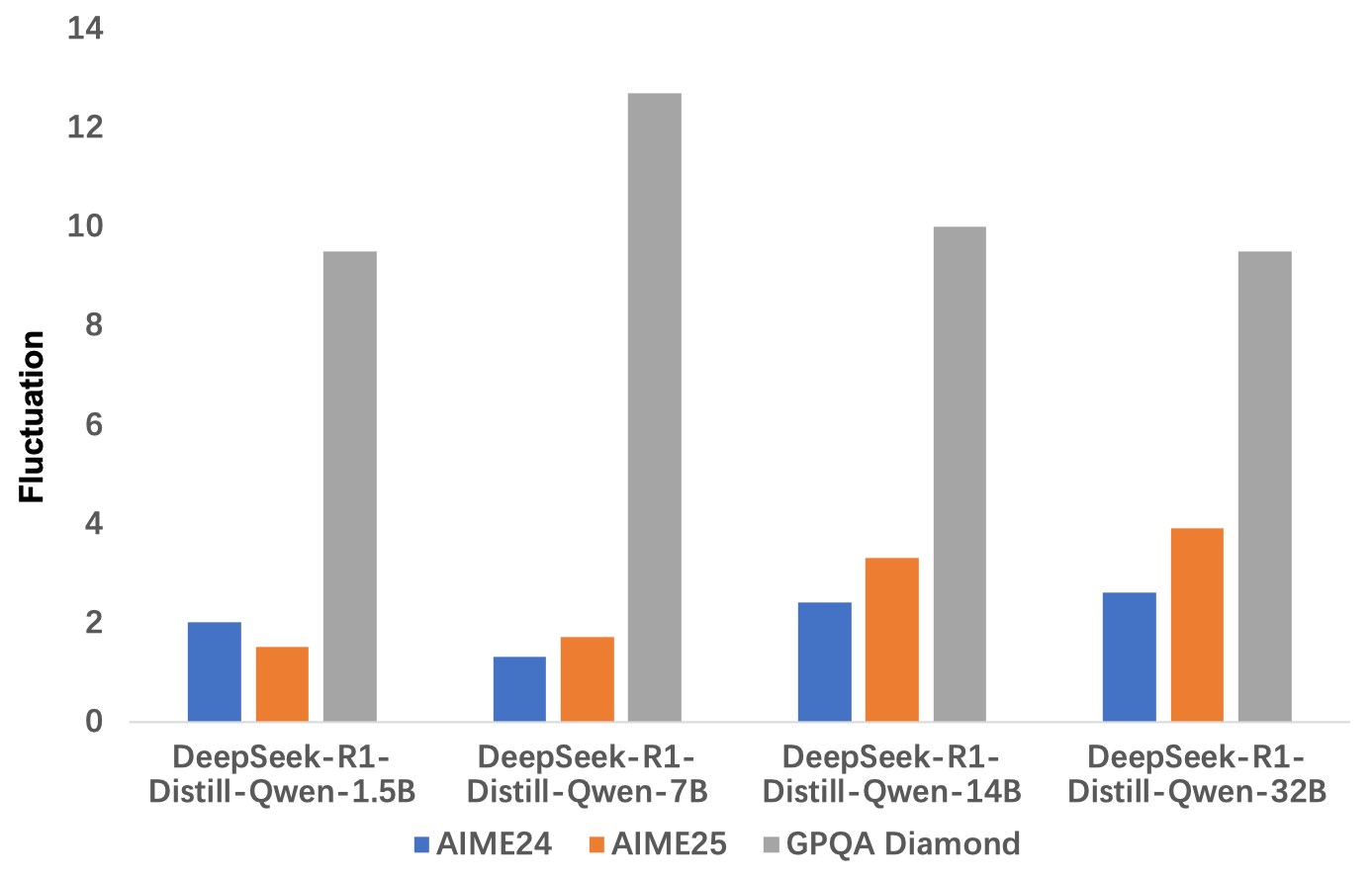}
    \caption{Score fluctuation ranges of the Deepseek-R1-Distill series models on relevant benchmarks under variations in subtle evaluation conditions that are often overlooked. The evaluation variables involved in this figure include: the version of the evaluation dataset, the relative position of the instruction, option bias and correct-answer bias in GPQA Diamond, and Tensor Parallelism settings.}
    \label{fig: figure1}
\end{figure}
            \end{minipage}
        \end{center}
    \end{strip}

    \twocolumn
    \section{Introduction}
\label{sec: intro}

Deepseek-R1-Distill series models \cite{deepseekai2025deepseekr1incentivizingreasoningcapability} have contributed significantly to the vitality of the open-source community and been widely praised by developers for their outstanding performance in various domains. We have also conducted practical applications based on the Deepseek-R1-Distill series and successfully reproduced their evaluation results on public benchmarks. In addition, we explored various other open-source inference models built on the Deepseek-R1-Distill series, as they frequently claim substantial improvements over the original models. However, we found that reproducing their benchmark results using the original evaluation code is challenging.

By examining the open-source evaluation scripts provided by the model developers, we gradually uncovered the underlying causes and identified several critical variables, such as the method of seed initialization and the version of evaluation data, that are frequently overlooked due to their subtlety and are inconsistently configured across evaluations.

In Section 2, we conduct comparative experiments to investigate the effects of these variables and discover that their impact far exceeds our initial expectations. This raises an important question: can configuration adjustments alone yield gains comparable to those achieved through model training? This observation leads us to further question the fairness and reliability of current evaluation practices.

Finally, in Section 3, we propose a rigorous and transparent paradigm for evaluating model performance, and provide detailed recommendations for improving evaluation practices. We hope that these issues will receive greater attention, as they directly affect developers’ decision-making in the open-source community, decisions that are often made at the cost of significant time and computational resources.

This study systematically reveals that subtle variations in evaluation design—the choice of N, random seed selection, benchmark dataset version, instruction placement, option bias and correct-answer bias in multiple-choice-questions-style benchmarks, and Tensor Parallelism settings—can lead to substantial fluctuations in benchmark scores for reasoning-focused LLMs such as the Deepseek-R1-Distill series and its derivatives. Our key findings demonstrate that:

\begin{itemize}[leftmargin=*]
    \item \textbf {Evaluation Conditions Critically Affect Results}: Minor changes can shift scores by several percentage points, undermining the reliability of model comparisons.
    \vspace{-0.5em}
    \item \textbf {Seed Initialization and N-Sampling}: Results are highly sensitive to seed choices; using larger N improves stability but requires thoughtful calibration per model and benchmark.
    \vspace{-0.5em}
    \item \textbf {Dataset Version Differences}: Inconsistencies in visual or formatting details can alter results by up to 3.9 percentage points, especially in math-related benchmarks like AIME.
    \vspace{-0.5em}
    \item \textbf {Option and Answer Biases}: In GPQA Diamond, option order and correct-answer placement can introduce performance swings exceeding 5 percentage points.
    \vspace{-0.5em}
    \item \textbf {Instruction placement and Tensor Parallelism}: While their individual impacts are relatively modest, they still affect reproducibility and should be documented.
\end{itemize}

Our analysis underscores that many claimed performance gains in open-source models are partially attributable to favorable evaluation setups rather than genuine model improvement. The widespread lack of transparency and standardized evaluation protocols leads to non-reproducible and potentially misleading results.

To address this, we recommend the community adopt a rigorous evaluation paradigm:

\begin{itemize}[leftmargin=*]
    \item Use dynamic seeds, document all settings transparently, and report confidence intervals rather than peak scores.
    \vspace{-0.5em}
    \item Calibrate N-sampling theoretically for stability, considering both model scale and benchmark characteristics.
    \vspace{-0.5em}
    \item Promote standardized evaluation frameworks to ensure fair and reproducible model comparisons.
\end{itemize}

By embracing these practices, the community can foster more reliable assessments of LLM reasoning capabilities and avoid overclaiming driven by evaluation artifacts.
    \section{Minor Variations, Major Fluctuations}

\begin{table*}[ht]
    \centering
    \small
    \begin{tabular}{llcccc}
        \toprule
        \multirow{2}{*}{Benchmark} & \multirow{2}{*}{Metric} & DeepSeek-R1- & DeepSeek-R1- & DeepSeek-R1- & DeepSeek-R1- \\
        & & Distill-Qwen-1.5B & Distill-Qwen-7B & Distill-Qwen-14B & Distill-Qwen-32B \\
        \midrule
        \multirow{3}{*}{AIME24} & Control Group & 31.2 & 54.4 & 69.2 & 71.8 \\
         & Repeated Trial & 31.2 & 55.8 & 69.1 & 71.8 \\ 
         & Fluctuation & 0.0 & 1.4 & 0.1 & 0.0 \\
        \midrule
        \multirow{3}{*}{AIME25} & Control Group & 23.7 & 40.0 & 52.0 & 56.6 \\
         & Repeated Trial & 23.7 & 40.4 & 51.9 & 56.6 \\ 
         & Fluctuation & 0.0 & 0.4 & 0.1 & 0.0 \\
        \midrule
        \multirow{3}{*}{GPQA Diamond} & Control Group & 40.3 & 54.7 & 61.3 & 67.4 \\
         & Repeated Trial & 40.2 & 55.0 & 62.0 & 67.5 \\
         & Fluctuation & 0.1 & 0.3 & 0.7 & 0.1 \\
        \bottomrule
    \end{tabular}
    \caption{A repeated trial is performed using the exact same configuration as the control group. The absolute difference between the two sets of results is used as the baseline fluctuation value for subsequent comparisons: (1) Control Group: Default configuration; Repeated Trial: Identical configuration rerun.
    (2) Unit: Percentage Points.}
    \label{tab: table1}
\end{table*}

We selected popular reasoning models on Hugging Face with more than 500 total downloads (as of April 26, 2025) as evaluation targets. These include:
\begin{itemize}[leftmargin=*]
    \item \textbf{32B scale}: DeepSeek-R1-Distill-Qwen-32B, QwQ-32B \cite{qwq32b, qwen2.5}, Skywork-OR1-32B-Preview \cite{he2025skywork, skywork-or1-2025}, TinyR1-32B-Preview \cite{tinyr132bpreview, tinyr1proj}
    \vspace{-0.5em}
    \item \textbf{14B scale}: DeepSeek-R1-Distill-Qwen-14B, DeepCoder-14B-Preview \cite{deepcoder2025}, Light-R1-14B-DS \cite{lightr1proj}
    \vspace{-0.5em}
    \item \textbf{7B scale}: DeepSeek-R1-Distill-Qwen-7B, Light-R1-7B-DS, Skywork-OR1-Math-7B
    \vspace{-0.5em}
    \item \textbf{1.5B scale}: DeepSeek-R1-Distill-Qwen-1.5B, DeepScaleR-1.5B-Preview \cite{deepscaler2025}, Open-RS1, Open-RS2, Open-RS3 \cite{dang2025reinforcementlearningreasoningsmall}, DeepCoder-1.5B-Preview, ZR1-1.5B \cite{zyphra2025ZR1}, OpenRS-GRPO, FastCuRL-1.5B-Preview \cite{fastcurl15bpreview}, STILL-3-1.5B-preview \cite{Slow_Thinking_with_LLMs_3_Preview, Slow_Thinking_with_LLMs_1, Slow_Thinking_with_LLMs_2}
\end{itemize}

Previous studies \cite{asoberlookatprogress, stresstestinggeneralization} have shown that inference parameters such as context length, temperature, top\_p, and top\_k can significantly affect results. However, this study does not focus on these extensively discussed factors. For these parameters, we follow the officially recommended values provided by the model publishers as of April 26, 2025. If such values are not available, we adopt the recommended parameters of the corresponding base model; otherwise, the defaults of vLLM \cite{kwon2023efficient} version 0.6.3 are used. Details of the selected models and their associated inference parameters are provided in the appendix~\ref{appendix:A.1}.

To enhance the clarity of our conclusions, unless otherwise specified, the experiments primarily focus on the 1.5B, 7B, 14B, and 32B variants of the Deepseek-R1-Distill series, evaluated on three benchmarks: AIME24, AIME25, and GPQA Diamond. The detailed results for all evaluated models are summarized in Section 2.8. In addition, we prioritize the presentation of key findings in the main text.

\subsection{Experimental Setup}

We investigated the effects of relevant variables through rigorous controlled experiments using the vLLM framework (version 0.6.3). The variables examined include: the choice of N (i.e., sampling N times and averaging results), seed initialization strategy, evaluation dataset version, the relative position of questions and instructions, option bias and correct-answer bias in GPQA Diamond \cite{gpqa}, and Tensor Parallelism settings. Unless otherwise specified, all experiments adopt the approach of N-samples  without explicitly setting a seed, which means each sample is replicated N times for inference, and the final result is obtained by averaging the pass@1 scores.

According to the vLLM implementation, if the SamplingParams method does not specify a seed, a dynamic seed (randomly generated per inference) is used. Therefore, the N-samples approach can be considered theoretically well-randomized.

Unless otherwise stated, all subsequent experiments adopt the following control group configuration:

\begin{itemize}[leftmargin=*]
    \item {N}: {64.}
    \vspace{-0.5em}
    \item {Seed}: {Dynamic seed.}
    \vspace{-0.5em}
    \item {Evaluation dataset version}: {\begin{itemize}[leftmargin=*]
    \vspace{-0.5em}
        \item {AIME24}: {simplescaling/aime24\_figures \cite{muennighoff2025s1simpletesttimescaling}.}
        \item {AIME25}: {simplescaling/aime25\_figures.}
        \item {GPQA Diamond}: {Idavidrein/gpqa.}
    \end{itemize}}
    \vspace{-0.5em}
    \item {Instruction position}: {Instruction placed after the question.}
    \vspace{-0.5em}
    \item {Option and answer bias in GPQA Diamond}: {Options ordered as (A → B → C → D), with the correct answer placed at A.}
    \vspace{-0.5em}
    \item {Tensor Parallelism setting}: {1 for model sizes no larger than 14B and 2 for 32B models.}
\end{itemize}

The software and hardware configurations that may influence the experimental results are detailed below:

\begin{itemize}[leftmargin=*]
    \item {GPU}: {NVIDIA H800 80GB.}
    \vspace{-0.5em}
    \item {Evaluation framework}: {verl \cite{sheng2024hybridflow}.}
\end{itemize}

To quantify the impact of each variable on model performance, we compute the magnitude of fluctuation as the absolute difference between the results of the experimental and control groups. Additionally, we conduct a repeated run of the control group using the exact same configuration to serve as the reference value for baseline fluctuation. The fluctuations observed are reported in Table \ref{tab: table1}. As noted above, N is set to 64 for this set of experiments.

\subsection{Average N}

As N increases, performing N independent inferences on the same question allows the evaluation result to better approximate the model’s true performance. In this section, we aim to systematically explore the extent to which the choice of N influences evaluation outcomes. We treat the result obtained with N = 64 as an approximate ground truth and examine the deviations of inference results at smaller N values from this reference.

\begin{figure}[H]
    \centering
    \includegraphics[width=0.96\linewidth]{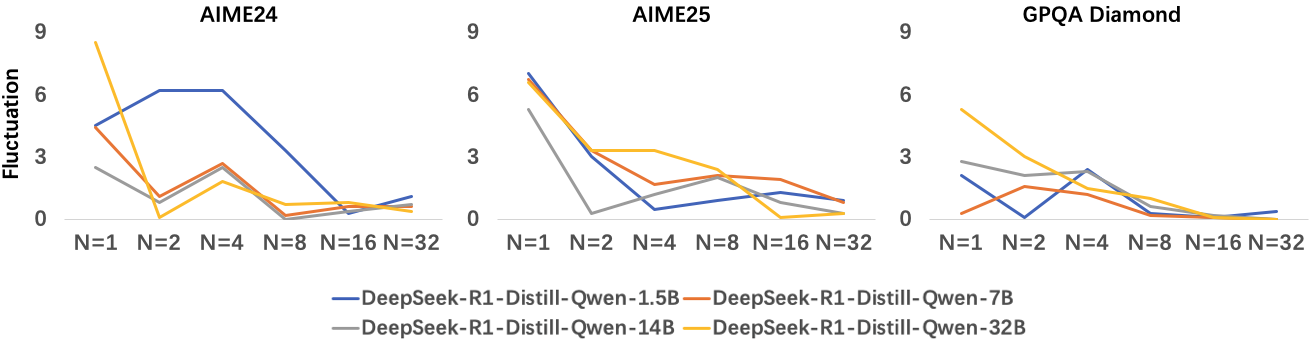}
    \caption{As the value of N increases, the fluctuation in performance across the four models on AIME24, AIME25, and GPQA Diamond gradually decreases, generally approaching within 1 percentage point at N = 32. It is also noteworthy that DeepSeek-R1-Distill-Qwen-1.5B exhibits the largest fluctuation, while GPQA Diamond, which features a larger sample size, exhibits relatively lower performance variance.}
    \label{fig: figure2}
\end{figure}

We define the fluctuation as the absolute deviation between an evaluation result and the approximate ground truth. The results are summarized in Figure~\ref{fig: figure2}. Across all four model variants, the fluctuations at N = 32 generally approach 1 percentage point. However, over 75\% of experiments still exhibit deviations beyond the baseline fluctuation range. Notably, performance fluctuation is not only influenced by the value of N, but also by the model size and the benchmark dataset used. For example, among the three benchmark tasks, the Deepseek-R1-Distill-Qwen-1.5B model shows the largest fluctuation, while the fluctuation is relatively smaller on GPQA Diamond, which contains a larger sample size. Detailed experimental results are provided in the appendix~\ref{appendix:A.2}.

This section demonstrates that simply altering the value of N is sufficient to cause substantial performance fluctuations across multiple benchmark tasks. Therefore, when reporting model performance, it is essential to explicitly specify the value of N used and to clarify its impact on result stability.

\subsection{Seed}

As mentioned previously, if the seed parameter is not explicitly specified in the SamplingParams method, the system automatically generates a dynamic seed for each inference. During our experiments, we observed that the result variations caused by different dynamic seeds far exceeded expectations, indicating that seed is one of the key factors affecting evaluation stability.

To further investigate the influence of seed on model outputs, we designed a fixed-seed N-times inference setup (denoted as 1-Seed-N), in which each sample is inferred N times using the same fixed seed, thereby obtaining stable outputs for benchmark evaluation.

\begin{table}[htbp]
\centering
    \small
    \begin{tabular}{cccccc}
        \toprule
        \multirow{1}{*}{Model Size} & N = 1 & N = 2 & N = 4 & N = 8 & N = 16 \\
        \midrule
        1.5B & 37 & 11 & 0 & 0 & 0 \\
        7B & 26 & 22 & 0 & 0 & 0 \\
        14B & 0 & 0 & 0 & 4 & 44 \\
        32B & 0 & 2 & 31 & 3 & 12 \\
        \bottomrule
    \end{tabular}
    \caption{Number of cases where evaluation results under a fixed seed stop changing at the minimum N for each model on the corresponding benchmarks. For example, the value 11 under N=2 for DeepSeek-R1-Distill-Qwen-1.5B indicates that in 11 out of the 16 1-Seed-N experiments, the evaluation results remained stable once N reached 2. This reflects the stability point of the model’s evaluation under a fixed seed on the given benchmarks.}
    \label{tab: table2}
\end{table}

In this setup, we randomly selected 16 integers from the range [0, 32767] as the seeds. The value of N was set to 16, based on our empirical findings across multiple experiments showing that over 70\% of evaluation results stabilize beyond N = 8 (see Table~\ref{tab: table2}). To balance output stability and computational cost, we ultimately chose N = 16, as it provided a good trade-off based on our observations.

We then measured model performance under each of the 16 fixed seeds across the selected benchmarks. As shown in Figure~\ref{fig: figure1}, across all models and benchmarks, the fluctuations caused by varying the seed are substantially greater than the baseline fluctuation, further confirming that seed is a critical variable affecting model stability. Notably, in some cases, small-scale models using specific seeds can match or even outperform larger models on certain benchmarks. This suggests that in the absence of standardized seed control, evaluation results may reflect misleading advantages. Detailed experimental results
are provided in the appendix~\ref{appendix:A.4}.

\begin{figure}[H]
    \centering
    \includegraphics[width=0.96\linewidth]{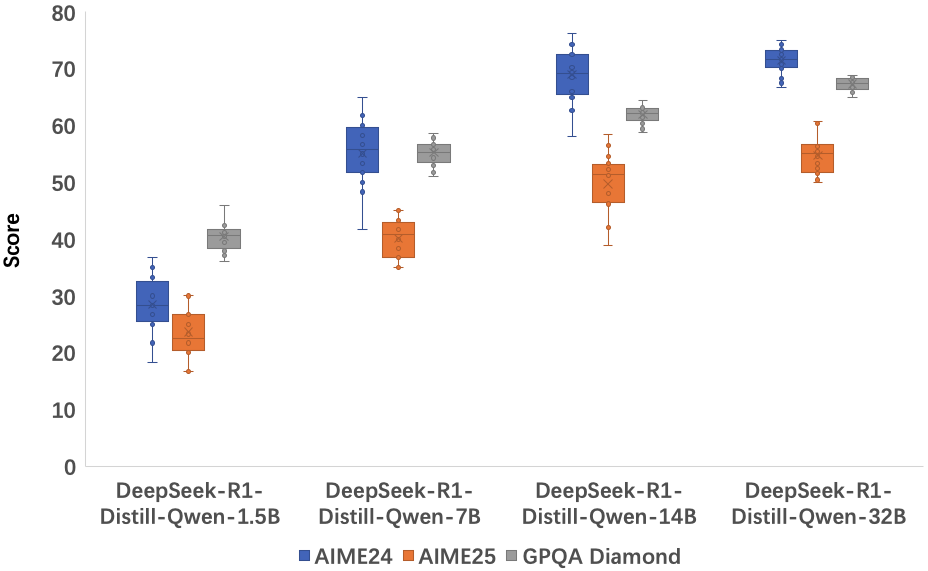}
    \caption{Under the 1-Seed-N setting, the evaluation result fluctuations caused by varying the seed are significantly greater than the baseline reference fluctuation, further confirming that the seed parameter is a critical factor influencing model stability.}
    \label{fig: figure3}
\end{figure}

This experiment clearly demonstrates the extreme sensitivity of model evaluation outcomes to the random seed setting, especially for tasks with small evaluation sample sizes.

\subsection{Evaluation Dataset Version}

While organizing benchmark data, we observed that multiple versions of the AIME evaluation datasets exist within the open-source community. The primary differences lie in how image-related information within the questions is processed. This raises an important question: to what extent do differences between dataset versions affect evaluation results? To investigate, we selected several representative datasets for comparative experiments. The results are summarized in Table~\ref{tab: table3}, and the dataset examples
are provided in the appendix~\ref{appendix:A.3}.

\begin{table*}[ht]
    \centering
    \small
    \begin{tabular}{lccc}
        \toprule
        Benchmark & Evaluation Dataset Version
         & Contains Image Description
         & Method \\
        \midrule
        \multirow{3}{*}{AIME24} & simplescaling/aime24\_figures (Control Group) & Yes & Asymptote \\
        & simplescaling/aime24\_nofigures & Omitted if Not Relevant to Solution
         & Asymptote \\
        & HuggingFaceH4/aime\_2024 \cite{huggingface2024aime} & No & - \\
        \midrule
        \multirow{3}{*}{AIME25} & simplescaling/aime25\_figures (Control Group) & Yes & Asymptote \\
        & simplescaling/aime25\_nofigures & Omitted if Not Relevant to Solution
         & Asymptote \\
        & yentinglin/aime\_2025 \cite{yentinglin2025aime} & Yes & Tikz \\
        \bottomrule
    \end{tabular}
    \caption{Overview of different versions of AIME evaluation datasets, categorized based on the handling of image information within the dataset.}
    \label{tab: table3}
\end{table*}

As previously mentioned, we treat simplescaling/aime24\_figures and simplescaling/aime25\_figures as the control group, and ensure that all other experimental variables are aligned with those of the control group to isolate the impact of dataset version on model performance.

\begin{figure}[H]
    \centering
    \includegraphics[width=0.96\linewidth]{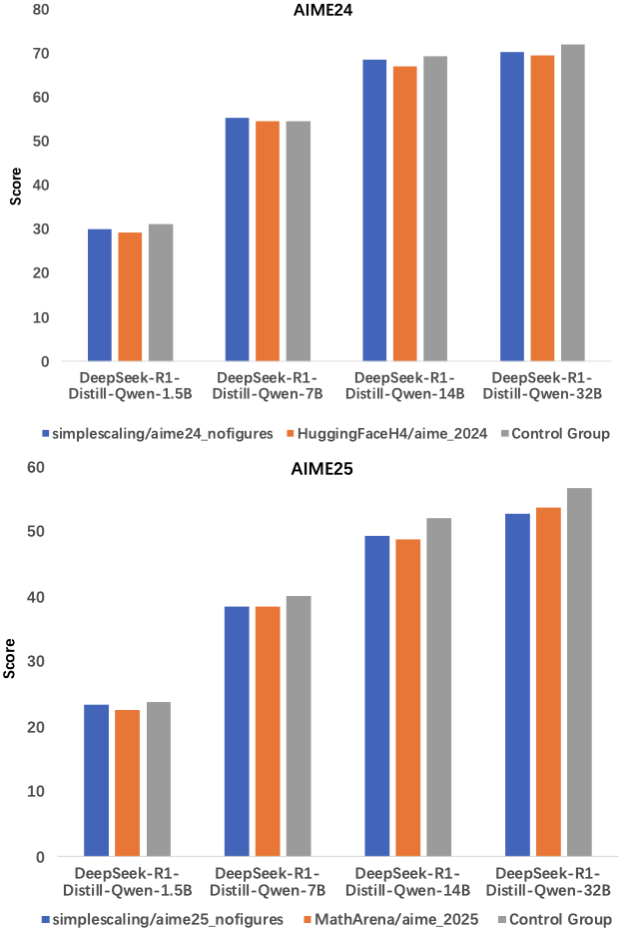}
    \caption{Fluctuations in benchmark scores across different versions of the AIME evaluation datasets. The control group—containing complete image information rendered using Asymptote—consistently outperforms other experimental groups.}
    \label{fig: figure4}
\end{figure}

The results in Figure~\ref{fig: figure4} indicate that performance variation for the same reasoning model across different versions of evaluation datasets is substantial, often exceeding the baseline reference fluctuation. The maximum observed discrepancy reached up to 3.9 percentage points. Moreover, in most cases, models achieved better evaluation scores on samples containing complete image information, suggesting that visual descriptions contribute positively to reasoning tasks.

These findings highlight the systematic impact of dataset version on model performance, which is particularly pronounced in math-related tasks involving visual information. Therefore, when reporting evaluation results, the specific dataset version used should be clearly documented to ensure evaluation fairness and reproducibility.

\subsection{Instruction Position}

An instruction refers to a predefined prompt included in the model input to help the model better understand the task and generate more accurate responses. In this experiment, we explicitly include the instruction in the input to examine how its position relative to the question affects model performance. For the AIME evaluation, we use the following instruction:

\vspace{0.5em}
\textit{Let's think step by step and output the final answer within \textbackslash\textbackslash boxed\{\}.}
\vspace{0.5em}

We investigate how the instruction’s relative position to the question affects model performance in AIME tasks, using the following experimental design:

\begin{itemize}[leftmargin=*]
    \item Experimental group: The instruction is placed before the question, separated by a space.
    \vspace{-0.5em}
    \item Control group: The instruction is placed after the question, also separated by a space.
\end{itemize}

\begin{figure}[H]
    \centering
    \includegraphics[width=0.96\linewidth]{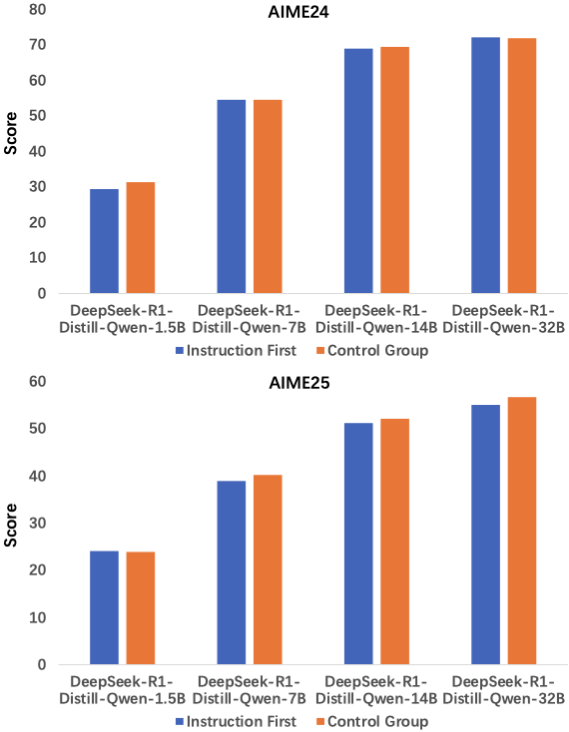}
    \caption{Compared to other variables, changes in benchmark scores caused by instruction position are relatively minor. However, they still introduce variations in evaluation stability. In certain model-benchmark combinations, placing the instruction before the question yields improved performance.}
    \label{fig: figure5}
\end{figure}

The experimental results are shown in Figure~\ref{fig: figure5}. The position of the instruction has a relatively minor impact on evaluation outcomes, with all variations falling below 2 percentage points. Notably, placing the instruction after the question generally yields better performance, potentially related to the input format used during the model’s training process—a hypothesis that warrants further investigation.

This experiment demonstrates that while instruction position does not cause drastic performance shifts, it can still affect the  evaluation stability. In some model and benchmark combinations, placing the instruction before the question yields slightly better results.

\begin{figure*}[t]
    \centering    \includegraphics[width=0.8\linewidth,height=0.25\textheight]{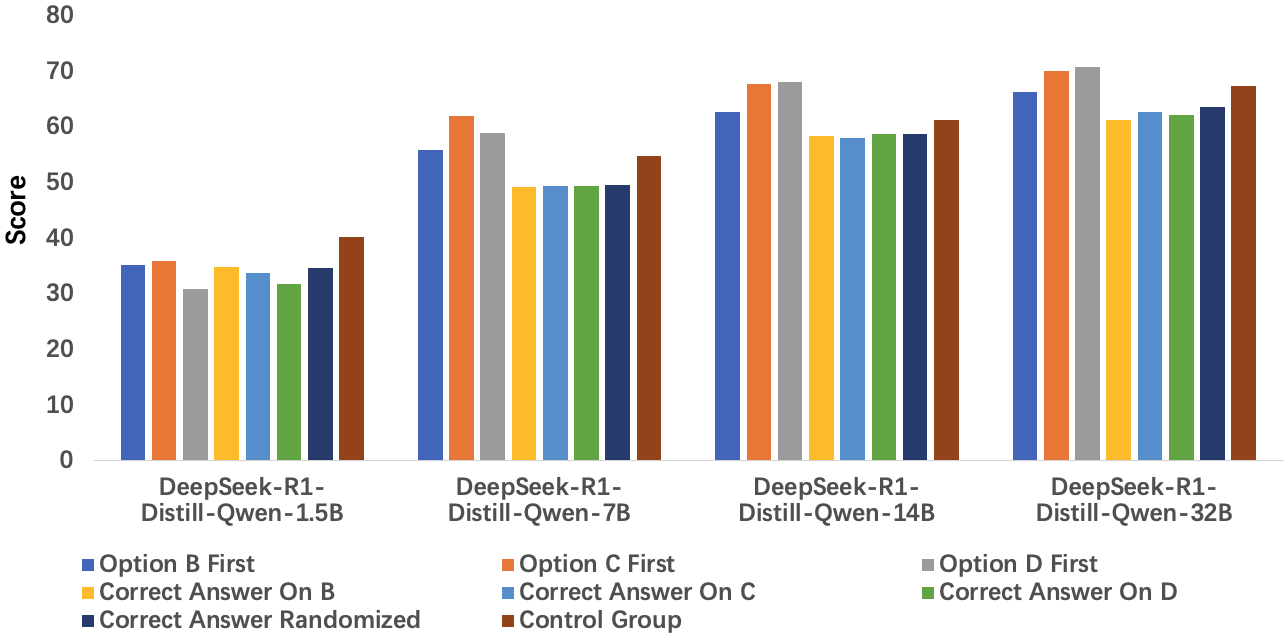}
    \caption{GPQA Diamond exhibits significant evaluation variability under option and correct answer biases. In particular, the randomized answer group consistently underperforms relative to the control group, suggesting that randomizing option order may undermine the model’s selection stability. Additionally, in the answer position bias experiments, the control group (with the correct answer placed immediately after the question) consistently outperforms all other groups, indicating that placing the correct answer early may enhance model performance.}
    \label{fig: figure6}
\end{figure*}

\subsection{Option and Answer Bias in GPQA Diamond}

Existing studies \cite{mcqswork} suggest that option order and the placement of the correct answer in multiple-choice questions (MCQs) can significantly influence model outputs. To verify the applicability of this phenomenon to reasoning models, we designed a series of controlled experiments based on the GPQA Diamond benchmark.

It is important to note that the official GPQA Diamond dataset does not fix option order or the correct answer’s position. However, some open-source evaluation frameworks default to placing the correct answer in option A. Whether this setup introduces systematic bias into evaluation results remains to be thoroughly validated. To address this, we constructed different permutations of evaluation samples and designed the following experimental groups:

\begin{itemize}[leftmargin=*]
    \item Control Group: Uses a fixed option order (A → B → C → D), with the correct answer always placed as option A.
    \vspace{-0.5em}
    \item Option Bias Group: The correct answer is always the first option, but the order is permuted as follows:
    \vspace{-0.5em}
    \begin{itemize}[leftmargin=*]
        \item (B → A → C → D),
        \vspace{-0.5em}
        \item (C → A → B → D),
        \vspace{-0.5em}
        \item (D → A → B → C).
        \vspace{-0.5em}
    \end{itemize}
    \item Answer Position Bias Group: The option order remains (A → B → C → D), but three subgroups are constructed where the correct answer is placed in option B, C, or D, respectively.
    \vspace{-0.5em}
    \item Randomized Group: Options are ordered as (A → B → C → D), but the correct answer is randomly placed in one of the four positions.
\end{itemize}

The results of these experiments are summarized in Figure~\ref{fig: figure6}.

The results show that although GPQA Diamond typically exhibits relatively low evaluation variability, the changes in option order and answer position in this experiment caused consistent and significant performance fluctuations—mostly above 5 percentage points. Specifically:

\begin{enumerate}
    \item For the majority of models, scores in the shuffled group were lower than those in the control group, indicating that randomizing option order may reduce model stability.
    \vspace{-0.5em}
    \item In the answer position bias experiments, the control group with the correct answer always in the first position consistently outperformed all other groups, suggesting that placing the correct answer directly after the question might improve model performance.
\end{enumerate}

This section highlights the significant impact of option order and correct answer position on model evaluation outcomes. It emphasizes the importance of standardizing the construction of MCQs. Without proper control, such biases may misrepresent a model’s true capabilities, undermining the fairness and reproducibility of results.

\subsection{Tensor Parallelism}

Although often overlooked, Tensor Parallelism (TP) configurations can impact the maximum number of output tokens a reasoning model can produce. This, in turn, determines the required maximum context length (max\_model\_len). Prior studies have shown that increasing max\_model\_len can improve model performance across multiple benchmarks, though it also demands more GPU memory. Increasing TP can help reduce memory pressure on a single GPU, thereby enabling a larger max\_model\_len. However, the internal communication and parallelization mechanisms involved in TP may introduce computational discrepancies, potentially affecting the final model output. Therefore, we conducted experiments to investigate how TP influences evaluation results.

\begin{figure}[H]
    \centering
    \includegraphics[width=0.96\linewidth]{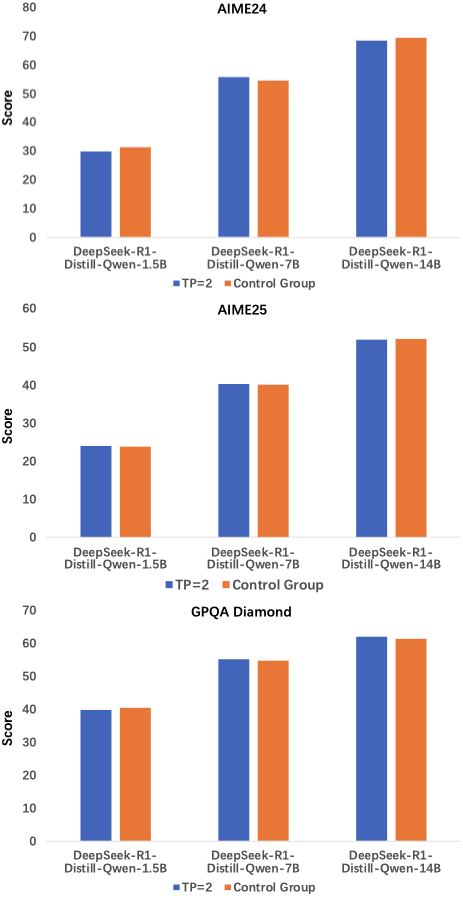}
    \caption{The impact of Tensor Parallelism (TP) variation on evaluation results is limited. However, for the sake of reproducibility, it is important to explicitly document the exact TP setting used.}
    \label{fig: figure7}
\end{figure}

In this section, we use DeepSeek-R1-Distill-Qwen-1.5B, DeepSeek-R1-Distill-Qwen-7B and DeepSeek-R1-Distill-Qwen-14B as the target models. These models have a default TP setting of 1. We compare their results to those obtained with TP set to 2.

The experimental results (Figure~\ref{fig: figure7}) show that changing the TP setting has limited impact on evaluation performance. All models exhibit fluctuations of less than 2 percentage points across benchmarks, with no consistent trend of improvement or decline. However, 67\% of the experimental groups exhibited fluctuation ranges that exceeded the baseline reference.

This experiment suggests that, under the current testing conditions, changing the TP parameter does not significantly affect model performance. However, for the sake of reproducibility—especially in memory-constrained environments—it is recommended that this parameter setting be explicitly specified.

\subsection{Performance of All Models}

We systematically applied the same methodology to evaluate all models under the relevant variables. The results are compiled in the appendix~\ref{appendix:A.5} for reference and reproducibility. The findings indicate that configuration settings—such as evaluation data versions and instruction placement—have a noticeable impact on the evaluation results of most models, confirming the generalizability of earlier observations. Overall, the evaluation results of reasoning models demonstrate consistent sensitivity to these factors. This sensitivity is consistently observed across models of varying sizes and diverse training origins, including the QwQ, Skywork, and OpenRS series.
    \begin{table*}[ht]
    \centering
    \small
    \begin{tabular}{llcccc}
    \toprule
    \multirow{2}{*}{Benchmark} & \multirow{2}{*}{Metric} & DeepSeek-R1- & DeepSeek-R1- & DeepSeek-R1- & DeepSeek-R1- \\
    & & Distill-Qwen-1.5B & Distill-Qwen-7B & Distill-Qwen-14B & Distill-Qwen-32B \\
    \midrule
    \multirow{4}{*}{AIME24} & Estimated Interval & 29.1±1 & 54.1±1 & 69.2±1 & 73.3±1 \\
     & Control Group & 31.2 & 54.4 & 69.2 & 71.8 \\
     & Official Result & 28.9 & 55.5 & 69.7 & 72.6 \\
     & Estimated N & 93 & 116 & 73 & 76 \\
    \midrule
    \multirow{4}{*}{AIME25} & Estimated Interval & 23.9±1 & 39.6±1 & 51.7±1 & 53.9±1 \\
     & Control Group & 23.7 & 40 & 52 & 56.6 \\
     & Official Result & - & - & - & - \\
     & Estimated N & 69 & 94 & 76 & 90 \\
    \midrule
    \multirow{4}{*}{GPQA Diamond} & Estimated Interval & 39.9±1 & 54.6±1 & 61.5±1 & 67.3±1 \\
     & Control Group & 40.3 & 54.7 & 61.3 & 67.4 \\
     & Official Result & 33.8 & 49.1 & 59.1 & 62.1 \\
     & Estimated N & 26 & 28 & 14 & 14 \\
    \bottomrule
\end{tabular}
\caption{Each experiment is evaluated by checking whether each control group result falls within the corresponding estimated 90\% confidence interval. It is evident that in evaluation methods that approximate the model’s true performance through increasing the number of repeated trials N, the appropriate value of N is dependent on the specific model and benchmark combination. Arbitrarily setting N to 16, 32, or 64 is not a rigorous or theoretically grounded practice. Note: (1) The control group represents the average result from 64 repeated trials. (2) The official results are as of April 25, 2025.}
\label{tab: table4}
\end{table*}

\section{Honestly Representing Model Performance is What You Truly Need}

Through the experiments above, we conducted detailed comparisons to examine how subtle changes in evaluation conditions can lead to fluctuations in benchmark results. For model users, benchmarks serve as a critical reference for model selection; for model developers, benchmarks provide a direct means of demonstrating model competitiveness. Unstandardized or non-transparent performance reporting can mislead members of the open-source community during model selection and deployment, resulting in significant waste of computational resources and time.

As researchers deeply involved in related experiments and reproducibility efforts, we recognize the severity of this issue. To address this, we propose a more rigorous paradigm in describing model performance—one that adheres to two fundamental principles: transparency and stability.

\subsection{Transparency}

Model evaluation results should extend beyond merely presenting comparative tables of benchmarks and model types. More importantly, all factors that could influence evaluation outcomes must be disclosed—including evaluation methodology, inference framework, parameter settings, data version, and hardware configuration.

Moreover, when conducting comparisons between models, the reproduced results of the baseline models under identical evaluation conditions should be provided, rather than merely citing their officially reported scores. This practice ensures the fairness and validity of model comparisons.

\subsection{Stability}

We contend that when publishing evaluation results, the primary emphasis should be placed on reporting the model’s stable performance on a benchmark, rather than its peak performance. Factors like random seed can cause significant variance and do not reflect the model’s true experience in real-world usage.

The use of the average-N approach for evaluating pass@1 is based on the idea of approximating a model’s actual performance by increasing the number of repeated trials. While many open-source projects indicate the N value used in their reported results, the rationale behind choosing a specific N is often unclear. For instance, in the case of DeepSeek-R1, all benchmark results are reported with N=64. However, benchmarks such as AIME24 (containing 30 samples) and GPQA Diamond (containing 198 samples) naturally require different N values to reach a stable measurement because of their differing sample sizes.

Therefore, we propose a theoretical basis to guide the selection of N. One can view a model’s stable performance on a benchmark as an approximation of the true distribution of its inference results. According to the Central Limit Theorem, for independent and identically distributed random variables, the sampling distribution of the standardized sample mean tends to approach a normal distribution—even if the original variables are not normally distributed. For normally distributed variables, it becomes possible to compute confidence intervals and confidence levels from the sample distribution, allowing us to establish a standard paradigm for evaluating the stability of model performance.

First, we need to define the desired form of stable results—for example, a confidence interval at 90\% confidence level, with a margin of error less than 2 percentage points. According to the formula:
\begin{equation}
    P\left( \left| \bar{x} - \mu \right| < \epsilon \right) \geq 1 - \alpha
\end{equation}
where $\epsilon$ denotes the error margin, $\alpha$ the significance level, $\bar{x}$ the sample mean, and $\mu$ the true mean of the model’s inference performance distribution on the benchmark.

From this, the confidence interval can be derived as:
\begin{equation}
    \bar{x} \pm z_{\alpha/2} \cdot \frac{s}{\sqrt{N}}
\end{equation}
where $s$ is the sample standard deviation, and $N$ is the number of repeated experiments. This leads to the derivation of the minimum required $N$:
\begin{equation}
    N \geq \left( \frac{z_{\alpha/2} \cdot s}{\epsilon} \right)^2
\end{equation}

Since the sample variance $s$ is computed based on repeated experiments, it is inherently unstable and varies with $N$. To address this issue, the ideal approach is to conduct a large number of experiments until s stabilizes. A more economical solution is to use an iterative procedure:

\begin{enumerate}
    \item Determine the initial experiment step size $N_0$, the significance level $\alpha$, and the error margin $\epsilon$, while fixing the evaluation environment.
    \vspace{-0.5em}
    \item Conduct $N_0$ repeated trials to compute the sample standard deviation $s_0$, then use the formula above to calculate the required number of trials $N_1$.
    \vspace{-0.5em}
    \item If $N_1 < N_0$, terminate the experiment and output the current result.
    \vspace{-0.5em}
    \item If $N_1 > N_0$, conduct $N_1 - N_0$ additional trials and repeat step 2 until the condition in step 3 is met.
\end{enumerate}

Based on the procedure above, we calculated the evaluation results and the final required N for the DeepSeek-R1-Distill-Qwen series models on AIME24, AIME25, and GPQA Diamond benchmarks under the conditions of $N_0 = 64$, $\alpha = 10\%$, and $\epsilon = 1$. The results are shown in Table~\ref{tab: table4}.

We observe whether the average result of each control group (based on N=64) falls within the corresponding estimated 90\% confidence interval. For GPQA Diamond, all 4 experiments largely satisfy this condition. However, for AIME24 and AIME25, 2 and 1 groups respectively fail to meet the criterion, further confirming that the appropriate $N$ value is benchmark-dependent. It is also noteworthy that the DeepSeek-R1-Distill-Qwen-32B model’s control group results for AIME24 and AIME25 both fall outside the estimated 90\% confidence interval, making it the most deviant model among the four. This suggests that the $N$ value is also model-dependent.

This chapter emphasizes the importance of transparency and stability as the two foundational principles in evaluating model performance. Relying solely on single-point metrics or “best-case” results often fails to reflect a model’s real-world reliability. We strongly recommend that the community adopt a more standardized and statistically grounded methodology to define a minimum experimental protocol for model evaluation, thereby enhancing fairness, comparability, and reproducibility.
    \section{Conclusion}

In this work, we systematically examined how subtle variations in evaluation conditions can lead to statistically significant fluctuations in benchmark scores when evaluating reasoning models. These variations include factors such as dataset versions, image description formats, instruction positions, option ordering, and Tensor Parallelism settings.

Our findings indicate that many popular benchmarks are sensitive to seemingly minor configuration changes. These changes can introduce biases, reduce reproducibility, and distort perceived model performance. Evaluation practices that prioritize best-case outcomes without considering variance may mislead both model users and developers.

We advocate for a more rigorous and standardized evaluation paradigm grounded in two core principles: transparency and stability. This includes fully disclosing evaluation settings and reporting not just peak performance, but also statistically supported stable performance (e.g., using confidence intervals and theoretically informed choices for N).

Only by fully embracing these principles can the open-source community ensure fairer model comparisons, prevent misleading claims, and foster the development of truly robust reasoning models.
    
    \onecolumn
    \bibliography{anthology,reference}
    \newpage
    \appendix
    \section{Appendix}
\vspace{-4em}

\subsection{All Models and Their Inference Parameters}
\label{appendix:A.1}

\vspace{-6em}

\begin{table}[H]
    \begin{tabular}{lcccccc}
        \toprule
        Model & temperature & max\_tokens & top\_p & top\_k & min\_p & presence\_penalty \\
        \midrule
        DeepSeek-R1-Distill-Qwen-32B & 0.6 & 32K & 0.95 & - & - & - \\
        QwQ-32B & 0.6 & 32K\footnotemark[1] & 0.95 & 30\footnotemark[2] & 0 & 1\footnotemark[3] \\
        Skywork-OR1-32B-Preview & 0.6 & 32K & 0.95 & - & - & - \\
        Tiny-R1-32B-preview & 0.6 & 32K & 0.95 & - & - & - \\
        DeepSeek-R1-Distill-Qwen-14B & 0.6 & 32K & 0.95 & - & - & - \\
        DeepCoder-14B-Preview & 0.6 & 64K & 0.95 & - & - & - \\
        Light-R1-14B-DS & 0.6 & 32K & 0.95 & - & - & - \\
        DeepSeek-R1-Distill-Qwen-7B & 0.6 & 32K & 0.95 & - & - & - \\
        Light-R1-7B-DS & 0.6 & 32K & 0.95 & - & - & - \\
        Skywork-OR1-Math-7B & 0.6 & 32K & 0.95 & - & - & - \\
        DeepSeek-R1-Distill-Qwen-1.5B & 0.6 & 32K & 0.95 & - & - & - \\
        DeepScaleR-1.5B-Preview & 0.6 & 32K & 0.95 & - & - & - \\
        Open-RS1\footnotemark[4] & 0.6 & 32K & 0.95 & - & - & - \\
        Open-RS2\footnotemark[4] & 0.6 & 32K & 0.95 & - & - & - \\
        Open-RS3\footnotemark[4] & 0.6 & 32K & 0.95 & - & - & - \\
        DeepCoder-1.5B-Preview & 0.6 & 32K & 0.95 & - & - & - \\
        ZR1-1.5B & 0.6 & 32K & 0.95 & - & - & - \\
        OpenRS-GRPO\footnotemark[4] & 0.6 & 32K & 0.95 & - & - & - \\
        FastCuRL-1.5B-Preview\footnotemark[5] & 0.6 & 32K & 1 & -1 & - & - \\
        STILL-3-1.5B-preview & 0.6 & 32K & 0.95 & - & - & - \\
        \bottomrule
    \end{tabular}
    \caption{Unless otherwise specified, the inference parameters are sourced from models' official Hugging Face homepages, with data current as of April 26, 2025.}
    \label{tab: table5}
\end{table}

\footnotetext[1]{A value of 32K was used to stay consistent with most of the other models since the official release did not provide the parameter.}
\footnotetext[2]{The officially recommended parameter is 20 to 40.}
\footnotetext[3]{The officially recommended parameter is 0 to 2.}
\footnotetext[4]{https://github.com/knoveleng/open-rs}
\footnotetext[5]{https://github.com/nick7nlp/FastCuRL}

\vspace{-4em}

\subsection{Detailed Experimental Results for the Section 2.2}
\label{appendix:A.2}

\vspace{-4em}

The following tables present the detailed statistical results corresponding to Figure~\ref{fig: figure2} in Section 2.2. The first column represents the number of inference repetitions N; columns 2 to 4 show the results of the N-th run, and the last three columns present the average results of the first N runs.

\begin{table}[H]
    \centering
    \small
    \begin{tabular}{ccccccc}
        \toprule
        Repetition Index & AIME24 & AIME25 & GPQA Diamond & Average AIME24 & Average AIME25 & Average GPQA Diamond \\
        \midrule
        1 & 26.7 & 16.7 & 42.4 & 26.7 & 16.7 & 42.4 \\
        2 & 23.3 & 36.7 & 38.4 & 25.0 & 26.7 & 40.4 \\
        3 & 23.3 & 23.3 & 34.9 & 24.4 & 25.6 & 38.6 \\
        4 & 26.7 & 20.0 & 35.9 & 25.0 & 24.2 & 37.9 \\
        5 & 40.0 & 23.3 & 47.0 & 28.0 & 24.0 & 39.7 \\
        6 & 30.0 & 26.7 & 41.9 & 28.3 & 24.4 & 40.1 \\
        7 & 30.0 & 26.7 & 39.9 & 28.6 & 24.8 & 40.0 \\
        8 & 23.3 & 23.3 & 39.4 & 27.9 & 24.6 & 40.0 \\
        9 & 33.3 & 26.7 & 41.9 & 28.5 & 24.8 & 40.2 \\
        10 & 26.7 & 30.0 & 38.9 & 28.3 & 25.3 & 40.0 \\
        11 & 36.7 & 23.3 & 36.9 & 29.1 & 25.2 & 39.8 \\
        12 & 43.3 & 20.0 & 43.9 & 30.3 & 24.7 & 40.1 \\
        13 & 33.3 & 26.7 & 41.4 & 30.5 & 24.9 & 40.2 \\
        14 & 30.0 & 20.0 & 40.4 & 30.5 & 24.5 & 40.2 \\
        15 & 40.0 & 26.7 & 42.4 & 31.1 & 24.7 & 40.4 \\
        16 & 36.7 & 30.0 & 38.4 & 31.5 & 25.0 & 40.2 \\
        17 & 36.7 & 23.3 & 41.4 & 31.8 & 24.9 & 40.3 \\
        18 & 43.3 & 16.7 & 34.3 & 32.4 & 24.4 & 40.0 \\
        19 & 20.0 & 20.0 & 40.4 & 31.8 & 24.2 & 40.0 \\
        20 & 30.0 & 30.0 & 40.9 & 31.7 & 24.5 & 40.0 \\
        21 & 40.0 & 30.0 & 39.9 & 32.1 & 24.8 & 40.0 \\
        22 & 30.0 & 20.0 & 38.9 & 32.0 & 24.5 & 40.0 \\
        23 & 30.0 & 20.0 & 42.9 & 31.9 & 24.3 & 40.1 \\
        24 & 43.3 & 26.7 & 40.4 & 32.4 & 24.4 & 40.1 \\
        25 & 26.7 & 23.3 & 36.9 & 32.1 & 24.4 & 40.0 \\
        26 & 36.7 & 23.3 & 38.4 & 32.3 & 24.4 & 39.9 \\
        27 & 33.3 & 26.7 & 39.9 & 32.3 & 24.4 & 39.9 \\
        28 & 40.0 & 26.7 & 39.9 & 32.6 & 24.5 & 39.9 \\
        29 & 36.7 & 26.7 & 46.0 & 32.8 & 24.6 & 40.1 \\
        30 & 26.7 & 23.3 & 42.4 & 32.6 & 24.6 & 40.2 \\
        31 & 23.3 & 33.3 & 35.9 & 32.3 & 24.8 & 40.1 \\
        32 & 33.3 & 16.7 & 34.9 & 32.3 & 24.6 & 39.9 \\
        33 & 23.3 & 26.7 & 45.0 & 32.0 & 24.6 & 40.1 \\
        34 & 30.0 & 30.0 & 37.9 & 32.0 & 24.8 & 40.0 \\
        35 & 33.3 & 16.7 & 42.4 & 32.0 & 24.6 & 40.1 \\
        36 & 33.3 & 30.0 & 41.9 & 32.0 & 24.7 & 40.1 \\
        37 & 33.3 & 23.3 & 45.0 & 32.1 & 24.7 & 40.3 \\
        38 & 23.3 & 20.0 & 39.4 & 31.8 & 24.6 & 40.2 \\
        39 & 36.7 & 20.0 & 37.4 & 32.0 & 24.4 & 40.2 \\
        40 & 36.7 & 20.0 & 41.4 & 32.1 & 24.3 & 40.2 \\
        41 & 33.3 & 16.7 & 36.9 & 32.1 & 24.1 & 40.1 \\
        42 & 36.7 & 16.7 & 40.4 & 32.2 & 24.0 & 40.1 \\
        43 & 30.0 & 33.3 & 41.9 & 32.2 & 24.2 & 40.2 \\
        44 & 30.0 & 30.0 & 41.9 & 32.1 & 24.3 & 40.2 \\
        45 & 20.0 & 30.0 & 35.4 & 31.9 & 24.4 & 40.1 \\
        46 & 30.0 & 26.7 & 39.9 & 31.8 & 24.5 & 40.1 \\
        47 & 23.3 & 20.0 & 40.4 & 31.6 & 24.4 & 40.1 \\
        48 & 30.0 & 26.7 & 42.9 & 31.6 & 24.4 & 40.2 \\
        49 & 26.7 & 20.0 & 40.4 & 31.5 & 24.4 & 40.2 \\
        50 & 26.7 & 23.3 & 44.4 & 31.4 & 24.3 & 40.2 \\
        51 & 36.7 & 23.3 & 34.9 & 31.5 & 24.3 & 40.1 \\
        52 & 36.7 & 13.3 & 42.9 & 31.6 & 24.1 & 40.2 \\
        53 & 26.7 & 30.0 & 48.0 & 31.5 & 24.2 & 40.3 \\
        54 & 36.7 & 13.3 & 38.4 & 31.6 & 24.0 & 40.3 \\
        55 & 33.3 & 23.3 & 39.4 & 31.6 & 24.0 & 40.3 \\
        56 & 23.3 & 23.3 & 40.9 & 31.5 & 24.0 & 40.3 \\
        57 & 33.3 & 13.3 & 37.9 & 31.5 & 23.8 & 40.3 \\
        58 & 33.3 & 26.7 & 46.0 & 31.6 & 23.9 & 40.4 \\
        59 & 30.0 & 23.3 & 41.4 & 31.5 & 23.8 & 40.4 \\
        60 & 23.3 & 23.3 & 43.4 & 31.4 & 23.8 & 40.4 \\
        61 & 26.7 & 26.7 & 38.4 & 31.3 & 23.9 & 40.4 \\
        62 & 23.3 & 23.3 & 39.9 & 31.2 & 23.9 & 40.4 \\
        63 & 30.0 & 20.0 & 35.4 & 31.2 & 23.8 & 40.3 \\
        64 & 33.3 & 20.0 & 40.4 & 31.2 & 23.7 & 40.3 \\
        \bottomrule
    \end{tabular}
    \caption{DeepSeek-R1-Distill-Qwen-1.5B}
    \label{tab: table6}
\end{table}

\begin{table}[H]
    \centering
    \small
    \begin{tabular}{ccccccc}
        \toprule
        Repetition Index & AIME24 & AIME25 & GPQA Diamond & Average AIME24 & Average AIME25 & Average GPQA Diamond \\
        \midrule
        1  & 50.0 & 33.3 & 55.0 & 50.0 & 33.3 & 55.0 \\
        2 & 56.7 & 40.0 & 57.6 & 53.3 & 36.7 & 56.3 \\
        3 & 46.7 & 43.3 & 47.0 & 51.1 & 38.9 & 53.2 \\
        4 & 53.3 & 36.7 & 54.5 & 51.7 & 38.3 & 53.5 \\
        5 & 63.3 & 40.0 & 58.6 & 54.0 & 38.7 & 54.5 \\
        6 & 46.7 & 40.0 & 57.1 & 52.8 & 38.9 & 55.0 \\
        7 & 60.0 & 53.3 & 54.5 & 53.8 & 41.0 & 54.9 \\
        8 & 60.0 & 50.0 & 54.5 & 54.6 & 42.1 & 54.9 \\
        9 & 50.0 & 53.3 & 57.6 & 54.1 & 43.3 & 55.2 \\
        10 & 46.7 & 46.7 & 53.0 & 53.3 & 43.7 & 55.0 \\
        11 & 53.3 & 36.7 & 52.0 & 53.3 & 43.0 & 54.7 \\
        12 & 53.3 & 40.0 & 56.6 & 53.3 & 42.8 & 54.8 \\
        13 & 56.7 & 43.3 & 59.6 & 53.6 & 42.8 & 55.2 \\
        14 & 63.3 & 40.0 & 53.0 & 54.3 & 42.6 & 55.1 \\
        15 & 56.7 & 43.3 & 55.0 & 54.4 & 42.7 & 55.1 \\
        16 & 63.3 & 30.0 & 47.5 & 55.0 & 41.9 & 54.6 \\
        17 & 56.7 & 40.0 & 51.0 & 55.1 & 41.8 & 54.4 \\
        18 & 60.0 & 36.7 & 53.0 & 55.4 & 41.5 & 54.3 \\
        19 & 56.7 & 40.0 & 51.5 & 55.4 & 41.4 & 54.1 \\
        20 & 56.7 & 46.7 & 55.6 & 55.5 & 41.7 & 54.2 \\
        21 & 50.0 & 33.3 & 53.5 & 55.2 & 41.3 & 54.2 \\
        22 & 56.7 & 36.7 & 55.0 & 55.3 & 41.1 & 54.2 \\
        23 & 70.0 & 46.7 & 55.0 & 55.9 & 41.3 & 54.3 \\
        24 & 56.7 & 36.7 & 52.0 & 56.0 & 41.1 & 54.2 \\
        25 & 50.0 & 43.3 & 57.1 & 55.7 & 41.2 & 54.3 \\
        26 & 50.0 & 50.0 & 53.5 & 55.5 & 41.5 & 54.3 \\
        27 & 63.3 & 36.7 & 57.6 & 55.8 & 41.4 & 54.4 \\
        28 & 46.7 & 40.0 & 59.6 & 55.5 & 41.3 & 54.6 \\
        29 & 53.3 & 46.7 & 53.5 & 55.4 & 41.5 & 54.5 \\
        30 & 46.7 & 33.3 & 54.5 & 55.1 & 41.2 & 54.5 \\
        31 & 56.7 & 36.7 & 63.6 & 55.2 & 41.1 & 54.8 \\
        32 & 50.0 & 33.3 & 51.0 & 55.0 & 40.8 & 54.7 \\
        33 & 50.0 & 43.3 & 57.1 & 54.8 & 40.9 & 54.8 \\
        34 & 53.3 & 40.0 & 52.0 & 54.8 & 40.9 & 54.7 \\
        35 & 63.3 & 33.3 & 58.6 & 55.0 & 40.7 & 54.8 \\
        36 & 56.7 & 40.0 & 55.6 & 55.1 & 40.6 & 54.8 \\
        37 & 46.7 & 43.3 & 57.6 & 54.9 & 40.7 & 54.9 \\
        38 & 63.3 & 33.3 & 56.6 & 55.1 & 40.5 & 54.9 \\
        39 & 56.7 & 40.0 & 52.5 & 55.1 & 40.5 & 54.9 \\
        40 & 60.0 & 36.7 & 55.0 & 55.3 & 40.4 & 54.9 \\
        41 & 60.0 & 26.7 & 50.5 & 55.4 & 40.1 & 54.8 \\
        42 & 53.3 & 46.7 & 55.6 & 55.3 & 40.2 & 54.8 \\
        43 & 66.7 & 46.7 & 52.0 & 55.6 & 40.4 & 54.7 \\
        44 & 63.3 & 40.0 & 53.0 & 55.8 & 40.4 & 54.7 \\
        45 & 56.7 & 30.0 & 55.0 & 55.8 & 40.1 & 54.7 \\
        46 & 50.0 & 33.3 & 54.0 & 55.7 & 40.0 & 54.7 \\
        47 & 43.3 & 40.0 & 53.0 & 55.4 & 40.0 & 54.7 \\
        48 & 66.7 & 40.0 & 48.5 & 55.6 & 40.0 & 54.5 \\
        49 & 56.7 & 36.7 & 52.5 & 55.6 & 39.9 & 54.5 \\
        50 & 56.7 & 43.3 & 55.0 & 55.7 & 40.0 & 54.5 \\
        51 & 43.3 & 43.3 & 56.1 & 55.4 & 40.1 & 54.5 \\
        52 & 50.0 & 50.0 & 56.1 & 55.3 & 40.3 & 54.6 \\
        53 & 43.3 & 33.3 & 58.6 & 55.1 & 40.1 & 54.6 \\
        54 & 50.0 & 33.3 & 52.5 & 55.0 & 40.0 & 54.6 \\
        55 & 53.3 & 40.0 & 53.5 & 55.0 & 40.0 & 54.6 \\
        56 & 53.3 & 43.3 & 55.6 & 54.9 & 40.1 & 54.6 \\
        57 & 36.7 & 33.3 & 54.0 & 54.6 & 39.9 & 54.6 \\
        58 & 50.0 & 50.0 & 52.0 & 54.5 & 40.1 & 54.5 \\
        59 & 50.0 & 36.7 & 49.0 & 54.5 & 40.1 & 54.4 \\
        60 & 56.7 & 36.7 & 56.1 & 54.5 & 40.0 & 54.5 \\
        61 & 56.7 & 40.0 & 57.1 & 54.5 & 40.0 & 54.5 \\
        62 & 53.3 & 46.7 & 54.5 & 54.5 & 40.1 & 54.5 \\
        63 & 43.3 & 30.0 & 63.1 & 54.3 & 39.9 & 54.7 \\
        64 & 56.7 & 43.3 & 59.1 & 54.4 & 40.0 & 54.7 \\
        \bottomrule
    \end{tabular}
    \caption{DeepSeek-R1-Distill-Qwen-7B}
    \label{tab: table7}
\end{table}

\begin{table}[H]
    \centering
    \small
    \begin{tabular}{ccccccc}
        \toprule
        Repetition Index & AIME24 & AIME25 & GPQA Diamond & Average AIME24 & Average AIME25 & Average GPQA Diamond \\
        \midrule
        1 & 66.7 & 46.7 & 64.1 & 66.7 & 46.7 & 64.1 \\
        2 & 73.3 & 56.7 & 62.6 & 70.0 & 51.7 & 63.4 \\
        3 & 56.7 & 50.0 & 62.6 & 65.6 & 51.1 & 63.1 \\
        4 & 70.0 & 50.0 & 65.2 & 66.7 & 50.8 & 63.6 \\
        5 & 66.7 & 33.3 & 64.7 & 66.7 & 47.3 & 63.8 \\
        6 & 70.0 & 53.3 & 62.1 & 67.2 & 48.3 & 63.6 \\
        7 & 80.0 & 53.3 & 55.6 & 69.0 & 49.0 & 62.4 \\
        8 & 70.0 & 56.7 & 58.1 & 69.2 & 50.0 & 61.9 \\
        9 & 70.0 & 46.7 & 57.6 & 69.3 & 49.6 & 61.4 \\
        10 & 70.0 & 53.3 & 59.1 & 69.3 & 50.0 & 61.2 \\
        11 & 70.0 & 53.3 & 62.1 & 69.4 & 50.3 & 61.3 \\
        12 & 73.3 & 53.3 & 64.1 & 69.7 & 50.6 & 61.5 \\
        13 & 66.7 & 53.3 & 63.6 & 69.5 & 50.8 & 61.7 \\
        14 & 63.3 & 53.3 & 60.1 & 69.0 & 51.0 & 61.5 \\
        15 & 63.3 & 56.7 & 59.6 & 68.7 & 51.3 & 61.4 \\
        16 & 70.0 & 50.0 & 62.6 & 68.8 & 51.2 & 61.5 \\
        17 & 70.0 & 60.0 & 62.1 & 68.8 & 51.8 & 61.5 \\
        18 & 76.7 & 56.7 & 61.6 & 69.3 & 52.0 & 61.5 \\
        19 & 60.0 & 40.0 & 62.1 & 68.8 & 51.4 & 61.6 \\
        20 & 70.0 & 50.0 & 59.6 & 68.8 & 51.3 & 61.5 \\
        21 & 73.3 & 40.0 & 59.6 & 69.0 & 50.8 & 61.4 \\
        22 & 73.3 & 50.0 & 60.1 & 69.2 & 50.8 & 61.3 \\
        23 & 80.0 & 56.7 & 63.6 & 69.7 & 51.0 & 61.4 \\
        24 & 76.7 & 53.3 & 61.1 & 70.0 & 51.1 & 61.4 \\
        25 & 70.0 & 50.0 & 62.1 & 70.0 & 51.1 & 61.4 \\
        26 & 76.7 & 53.3 & 60.1 & 70.3 & 51.2 & 61.4 \\
        27 & 66.7 & 53.3 & 60.6 & 70.1 & 51.2 & 61.4 \\
        28 & 73.3 & 50.0 & 60.6 & 70.2 & 51.2 & 61.3 \\
        29 & 70.0 & 56.7 & 61.1 & 70.2 & 51.4 & 61.3 \\
        30 & 60.0 & 50.0 & 62.6 & 69.9 & 51.3 & 61.4 \\
        31 & 76.7 & 53.3 & 60.1 & 70.1 & 51.4 & 61.3 \\
        32 & 63.3 & 60.0 & 60.1 & 69.9 & 51.7 & 61.3 \\
        33 & 66.7 & 50.0 & 62.6 & 69.8 & 51.6 & 61.3 \\
        34 & 60.0 & 53.3 & 59.6 & 69.5 & 51.7 & 61.3 \\
        35 & 70.0 & 50.0 & 58.6 & 69.5 & 51.6 & 61.2 \\
        36 & 66.7 & 50.0 & 57.1 & 69.4 & 51.6 & 61.1 \\
        37 & 60.0 & 50.0 & 61.1 & 69.2 & 51.5 & 61.1 \\
        38 & 73.3 & 53.3 & 57.1 & 69.3 & 51.6 & 61.0 \\
        39 & 76.7 & 53.3 & 59.1 & 69.5 & 51.6 & 60.9 \\
        40 & 66.7 & 43.3 & 60.1 & 69.4 & 51.4 & 60.9 \\
        41 & 66.7 & 50.0 & 62.6 & 69.4 & 51.4 & 61.0 \\
        42 & 76.7 & 63.3 & 61.1 & 69.5 & 51.7 & 61.0 \\
        43 & 73.3 & 56.7 & 60.1 & 69.6 & 51.8 & 60.9 \\
        44 & 66.7 & 56.7 & 64.7 & 69.5 & 51.9 & 61.0 \\
        45 & 70.0 & 43.3 & 63.1 & 69.6 & 51.7 & 61.1 \\
        46 & 73.3 & 53.3 & 64.1 & 69.6 & 51.7 & 61.1 \\
        47 & 70.0 & 50.0 & 64.7 & 69.6 & 51.7 & 61.2 \\
        48 & 66.7 & 60.0 & 60.1 & 69.6 & 51.9 & 61.2 \\
        49 & 66.7 & 50.0 & 64.7 & 69.5 & 51.8 & 61.3 \\
        50 & 60.0 & 43.3 & 62.1 & 69.3 & 51.7 & 61.3 \\
        51 & 73.3 & 46.7 & 63.1 & 69.4 & 51.6 & 61.3 \\
        52 & 73.3 & 46.7 & 62.1 & 69.5 & 51.5 & 61.3 \\
        53 & 60.0 & 46.7 & 58.1 & 69.3 & 51.4 & 61.3 \\
        54 & 63.3 & 53.3 & 65.2 & 69.2 & 51.4 & 61.3 \\
        55 & 73.3 & 53.3 & 57.6 & 69.3 & 51.5 & 61.3 \\
        56 & 63.3 & 56.7 & 63.1 & 69.2 & 51.5 & 61.3 \\
        57 & 63.3 & 56.7 & 61.1 & 69.1 & 51.6 & 61.3 \\
        58 & 66.7 & 46.7 & 63.1 & 69.0 & 51.6 & 61.3 \\
        59 & 70.0 & 53.3 & 63.6 & 69.0 & 51.6 & 61.4 \\
        60 & 70.0 & 56.7 & 61.6 & 69.1 & 51.7 & 61.4 \\
        61 & 70.0 & 60.0 & 61.1 & 69.1 & 51.8 & 61.4 \\
        62 & 70.0 & 53.3 & 60.1 & 69.1 & 51.8 & 61.3 \\
        63 & 73.3 & 56.7 & 62.6 & 69.2 & 51.9 & 61.4 \\
        64 & 70.0 & 56.7 & 58.6 & 69.2 & 52.0 & 61.3 \\
        \bottomrule
    \end{tabular}
    \caption{DeepSeek-R1-Distill-Qwen-14B}
    \label{tab: table8}
\end{table}

\begin{table}[H]
    \centering
    \small
    \begin{tabular}{ccccccc}
        \toprule
        Repetition Index & AIME24 & AIME25 & GPQA Diamond & Average AIME24 & Average AIME25 & Average GPQA Diamond \\
        \midrule
        1 & 63.3 & 50.0 & 62.1 & 63.3 & 50.0 & 62.1 \\
        2 & 80.0 & 56.7 & 66.7 & 71.7 & 53.3 & 64.4 \\
        3 & 76.7 & 43.3 & 67.7 & 73.3 & 50.0 & 65.5 \\
        4 & 60.0 & 63.3 & 67.2 & 70.0 & 53.3 & 65.9 \\
        5 & 80.0 & 53.3 & 66.7 & 72.0 & 53.3 & 66.1 \\
        6 & 63.3 & 56.7 & 67.2 & 70.6 & 53.9 & 66.2 \\
        7 & 76.7 & 63.3 & 64.7 & 71.4 & 55.2 & 66.0 \\
        8 & 80.0 & 46.7 & 69.2 & 72.5 & 54.2 & 66.4 \\
        9 & 66.7 & 70.0 & 68.7 & 71.9 & 55.9 & 66.7 \\
        10 & 70.0 & 56.7 & 69.7 & 71.7 & 56.0 & 67.0 \\
        11 & 73.3 & 60.0 & 65.2 & 71.8 & 56.4 & 66.8 \\
        12 & 76.7 & 66.7 & 69.7 & 72.2 & 57.2 & 67.0 \\
        13 & 63.3 & 53.3 & 68.7 & 71.5 & 56.9 & 67.2 \\
        14 & 66.7 & 53.3 & 68.7 & 71.2 & 56.7 & 67.3 \\
        15 & 63.3 & 50.0 & 69.2 & 70.7 & 56.2 & 67.4 \\
        16 & 76.7 & 63.3 & 68.2 & 71.0 & 56.7 & 67.5 \\
        17 & 66.7 & 56.7 & 67.2 & 70.8 & 56.7 & 67.4 \\
        18 & 73.3 & 50.0 & 68.7 & 70.9 & 56.3 & 67.5 \\
        19 & 66.7 & 60.0 & 75.8 & 70.7 & 56.5 & 67.9 \\
        20 & 63.3 & 56.7 & 66.7 & 70.3 & 56.5 & 67.9 \\
        21 & 70.0 & 53.3 & 65.7 & 70.3 & 56.3 & 67.8 \\
        22 & 70.0 & 60.0 & 65.2 & 70.3 & 56.5 & 67.7 \\
        23 & 73.3 & 53.3 & 65.7 & 70.4 & 56.4 & 67.6 \\
        24 & 70.0 & 60.0 & 66.2 & 70.4 & 56.5 & 67.5 \\
        25 & 70.0 & 53.3 & 65.7 & 70.4 & 56.4 & 67.4 \\
        26 & 76.7 & 56.7 & 67.2 & 70.6 & 56.4 & 67.4 \\
        27 & 73.3 & 60.0 & 66.7 & 70.7 & 56.5 & 67.4 \\
        28 & 70.0 & 56.7 & 66.7 & 70.7 & 56.5 & 67.4 \\
        29 & 80.0 & 60.0 & 69.7 & 71.0 & 56.7 & 67.5 \\
        30 & 70.0 & 53.3 & 63.6 & 71.0 & 56.6 & 67.3 \\
        31 & 73.3 & 63.3 & 67.2 & 71.1 & 56.8 & 67.3 \\
        32 & 80.0 & 60.0 & 70.2 & 71.4 & 56.9 & 67.4 \\
        33 & 70.0 & 50.0 & 66.7 & 71.3 & 56.7 & 67.4 \\
        34 & 73.3 & 63.3 & 65.7 & 71.4 & 56.9 & 67.3 \\
        35 & 70.0 & 63.3 & 68.2 & 71.3 & 57.0 & 67.4 \\
        36 & 70.0 & 60.0 & 66.7 & 71.3 & 57.1 & 67.3 \\
        37 & 76.7 & 50.0 & 69.7 & 71.4 & 56.9 & 67.4 \\
        38 & 73.3 & 56.7 & 61.1 & 71.5 & 56.9 & 67.2 \\
        39 & 80.0 & 60.0 & 68.7 & 71.7 & 57.0 & 67.3 \\
        40 & 76.7 & 53.3 & 66.7 & 71.8 & 56.9 & 67.3 \\
        41 & 63.3 & 63.3 & 66.2 & 71.6 & 57.1 & 67.2 \\
        42 & 76.7 & 60.0 & 69.2 & 71.7 & 57.1 & 67.3 \\
        43 & 70.0 & 60.0 & 70.7 & 71.7 & 57.2 & 67.4 \\
        44 & 66.7 & 53.3 & 68.7 & 71.6 & 57.1 & 67.4 \\
        45 & 70.0 & 60.0 & 72.7 & 71.6 & 57.2 & 67.5 \\
        46 & 73.3 & 66.7 & 64.7 & 71.6 & 57.4 & 67.4 \\
        47 & 66.7 & 56.7 & 69.7 & 71.5 & 57.4 & 67.5 \\
        48 & 66.7 & 60.0 & 65.2 & 71.4 & 57.4 & 67.4 \\
        49 & 66.7 & 63.3 & 67.7 & 71.3 & 57.6 & 67.5 \\
        50 & 73.3 & 53.3 & 66.2 & 71.3 & 57.5 & 67.4 \\
        51 & 73.3 & 50.0 & 67.2 & 71.4 & 57.3 & 67.4 \\
        52 & 70.0 & 50.0 & 68.7 & 71.3 & 57.2 & 67.4 \\
        53 & 70.0 & 56.7 & 66.7 & 71.3 & 57.2 & 67.4 \\
        54 & 70.0 & 50.0 & 69.2 & 71.3 & 57.0 & 67.5 \\
        55 & 83.3 & 60.0 & 67.7 & 71.5 & 57.1 & 67.5 \\
        56 & 76.7 & 53.3 & 66.2 & 71.6 & 57.0 & 67.4 \\
        57 & 76.7 & 63.3 & 68.2 & 71.7 & 57.1 & 67.5 \\
        58 & 76.7 & 53.3 & 68.7 & 71.8 & 57.1 & 67.5 \\
        59 & 70.0 & 53.3 & 65.2 & 71.8 & 57.0 & 67.4 \\
        60 & 73.3 & 63.3 & 68.2 & 71.8 & 57.1 & 67.5 \\
        61 & 66.7 & 43.3 & 65.7 & 71.7 & 56.9 & 67.4 \\
        62 & 80.0 & 43.3 & 65.7 & 71.8 & 56.7 & 67.4 \\
        63 & 76.7 & 56.7 & 69.2 & 71.9 & 56.7 & 67.4 \\
        64 & 66.7 & 53.3 & 67.7 & 71.8 & 56.6 & 67.4 \\
        \bottomrule
    \end{tabular}
    \caption{DeepSeek-R1-Distill-Qwen-32B}
    \label{tab: table9}
\end{table}

\subsection{Examples of Different Evaluation Dataset Versions in Section 2.4}
\label{appendix:A.3}

The following examples illustrate different versions of AIME evaluation datasets, highlighting variations in how image-related information is presented across datasets.

\vspace{2em}

\begin{itemize}[leftmargin=*]
    \item \textbf{simplescaling/aime24\_figures}: {Eight circles of radius \$34\$ are sequentially tangent, and two of the circles are tangent to \$AB\$ and \$BC\$ of triangle \$ABC\$, respectively. \$2024\$ circles of radius \$1\$ can be arranged in the same manner. The inradius of triangle \$ABC\$ can be expressed as \$\textbackslash frac\{m\}\{n\}\$, where \$m\$ and \$n\$ are relatively prime positive integers. Find \$m+n\$. [asy] pair A = (2,1); pair B = (0,0); pair C = (3,0); dot(A\textasciicircum\textasciicircum B\textasciicircum\textasciicircum C); label("\$A\$", A, N); label("\$B\$", B, S); label("\$C\$", C, S); draw(A--B--C--cycle); for(real i=0.62; i<2.7; i+=0.29)\{ draw(circle((i,0.145), 0.145)); \} [/asy]}
    \item \textbf{simplescaling/aime24\_nofigures}: {Eight circles of radius \$34\$ are sequentially tangent, and two of the circles are tangent to \$AB\$ and \$BC\$ of triangle \$ABC\$, respectively. \$2024\$ circles of radius \$1\$ can be arranged in the same manner. The inradius of triangle \$ABC\$ can be expressed as \$\textbackslash frac\{m\}\{n\}\$, where \$m\$ and \$n\$ are relatively prime positive integers. Find \$m+n\$.}
    \item \textbf{HuggingFaceH4/aime\_2024}: {Eight circles of radius \$34\$ are sequentially tangent, and two of the circles are tangent to \$AB\$ and \$BC\$ of triangle \$ABC\$, respectively. \$2024\$ circles of radius \$1\$ can be arranged in the same manner. The inradius of triangle \$ABC\$ can be expressed as \$\textbackslash frac\{m\}\{n\}\$, where \$m\$ and \$n\$ are relatively prime positive integers. Find \$m+n\$.}
    \item \textbf{simplescaling/aime25\_figures}: {Four unit squares form a \$2 \textbackslash times 2\$ grid. Each of the \$12\$ unit line segments forming the sides of the squares is colored either red or blue in such a way that each unit square has \$2\$ red sides and \$2\$ blue sides. One example is shown below (red is solid, blue is dashed). Find the number of such colorings. [asy] size(4cm); defaultpen(linewidth(1.2)); draw((0, 0) -- (2, 0) -- (2, 1)); draw((0, 1) -- (1, 1) -- (1, 2) -- (2,2)); draw((0, 0) -- (0, 1), dotted); draw((1, 0) -- (1, 1) -- (2, 1) -- (2, 2), dotted); draw((0, 1) -- (0, 2) -- (1, 2), dotted); [/asy]}
    \item \textbf{simplescaling/aime25\_nofigures}: {Four unit squares form a \$2 \textbackslash times 2\$ grid. Each of the \$12\$ unit line segments forming the sides of the squares is colored either red or blue in such a way that each unit square has \$2\$ red sides and \$2\$ blue sides. Find the number of such colorings.}
    \item \textbf{yentinglin/aime\_2025}: {Four unit squares form a \$2\textbackslash\textbackslash times 2\$ grid. Each of the \$12\$ unit line segments forming the sides of the squares is colored either red or blue in such a way that each unit square has \$2\$ red sides and \$2\$ blue sides. One example is shown below (red is solid, blue is dashed). Find the number of such colorings.\textbackslash n\textbackslash n\textbackslash n\textbackslash n\textbackslash\textbackslash begin\{tikzpicture\}\textbackslash n\textbackslash n    \textbackslash\textbackslash foreach \textbackslash\textbackslash x in \{0,1\} \{\textbackslash n\textbackslash n        \textbackslash\textbackslash foreach \textbackslash\textbackslash y in \{0,1\} \{\textbackslash n\textbackslash n            \textbackslash\textbackslash draw[dashed, blue, very thick] (\textbackslash \textbackslash x, \textbackslash \textbackslash y) rectangle ++(1,1);\textbackslash n\textbackslash n        \}\textbackslash n\textbackslash n    \}\textbackslash n\textbackslash n    \textbackslash n\textbackslash n    \textbackslash\textbackslash draw[red, very thick] (1,2) -- (2,2);\textbackslash n\textbackslash n    \textbackslash\textbackslash draw[red, very thick] (1,1) -- (1,2);\textbackslash n\textbackslash n    \textbackslash \textbackslash draw[red, very thick] (0,1) -- (1,1);\textbackslash n\textbackslash n    \textbackslash\textbackslash draw[red, very thick] (2,0) -- (2,1);\textbackslash n\textbackslash n    \textbackslash\textbackslash draw[red, very thick] (1,0) -- (2,0);\textbackslash n\textbackslash n    \textbackslash\textbackslash draw[red, very thick] (0,0) -- (1,0);\textbackslash n\textbackslash n\textbackslash\textbackslash end\{tikzpicture\}}
\end{itemize}

\subsection{Detailed Experimental Results for the Section 2.3}
\label{appendix:A.4}

We present the detailed statistical results behind Figure~\ref{fig: figure3} in Section 2.3 in the following tables. The last three columns present the average results of the model under the fixed seed across 16 runs.

\begin{table}[H]
    \centering
    \small
    \begin{tabular}{llccc}
        \toprule
        Model & Fixed Seed & AIME24 & AIME25 & GPQA Diamond \\
        \midrule
        \multirow{16}{*}{DeepSeek-R1-Distill-Qwen-1.5B
} & 28354 & 35.0 & 25.0 & 40.7 \\
         & 24624 & 28.3 & 21.7 & 37.9 \\
         & 26486 & 26.7 & 30.0 & 40.7 \\
         & 726 & 18.3 & 26.7 & 40.4 \\
         & 18595 & 25.0 & 26.7 & 41.2 \\
         & 24912 & 30.0 & 20.0 & 37.1 \\
         & 4002 & 26.7 & 30.0 & 41.9 \\
         & 13263 & 28.3 & 23.3 & 37.9 \\
         & 23242 & 36.7 & 20.0 & 46.0 \\
         & 13839 & 28.3 & 30.0 & 39.4 \\
         & 30367 & 33.3 & 16.7 & 41.4 \\
         & 2904 & 21.7 & 26.7 & 40.4 \\
         & 7716 & 25.0 & 21.7 & 41.4 \\
         & 8832 & 28.3 & 16.7 & 36.1 \\
         & 20000 & 33.3 & 21.7 & 42.4 \\
         & 15724 & 30.0 & 21.7 & 41.9 \\
        \midrule
        \multirow{16}{*}{DeepSeek-R1-Distill-Qwen-7B
} & 28354 & 61.7 & 35.0 & 53.0 \\
         & 24624 & 58.3 & 36.7 & 51.8 \\
         & 26486 & 60.0 & 41.7 & 58.6 \\
         & 726 & 55.0 & 41.7 & 57.8 \\
         & 18595 & 41.7 & 40.0 & 54.8 \\
         & 24912 & 53.3 & 43.3 & 54.8 \\
         & 4002 & 51.7 & 43.3 & 58.3 \\
         & 13263 & 48.3 & 38.3 & 56.6 \\
         & 23242 & 51.7 & 45.0 & 56.1 \\
         & 13839 & 50.0 & 40.0 & 53.3 \\
         & 30367 & 51.7 & 41.7 & 51.0 \\
         & 2904 & 58.3 & 36.7 & 56.1 \\
         & 7716 & 65.0 & 41.7 & 55.8 \\
         & 8832 & 56.7 & 45.0 & 56.8 \\
         & 20000 & 60.0 & 35.0 & 54.8 \\
         & 15724 & 58.3 & 36.7 & 54.3 \\
        \midrule
        \multirow{16}{*}{DeepSeek-R1-Distill-Qwen-14B
} & 28354 & 74.2 & 56.5 & 63.1 \\
         & 24624 & 65.2 & 42.1 & 61.9 \\
         & 26486 & 70.4 & 53.3 & 62.2 \\
         & 726 & 70.2 & 58.5 & 58.8 \\
         & 18595 & 69.0 & 51.2 & 61.4 \\
         & 24912 & 69.4 & 48.1 & 62.8 \\
         & 4002 & 74.2 & 52.3 & 63.3 \\
         & 13263 & 76.2 & 54.6 & 62.1 \\
         & 23242 & 66.0 & 52.5 & 62.1 \\
         & 13839 & 68.5 & 46.2 & 59.4 \\
         & 30367 & 68.8 & 52.3 & 60.4 \\
         & 2904 & 65.0 & 51.7 & 64.4 \\
         & 7716 & 62.7 & 42.1 & 61.9 \\
         & 8832 & 72.5 & 38.8 & 63.1 \\
         & 20000 & 58.1 & 46.9 & 62.7 \\
         & 15724 & 72.5 & 48.3 & 60.8 \\
        \midrule
        \multirow{16}{*}{DeepSeek-R1-Distill-Qwen-32B
} & 28354 & 70.8 & 50.0 & 66.2 \\
         & 24624 & 74.2 & 54.6 & 68.4 \\
         & 26486 & 66.7 & 55.8 & 65.0 \\
         & 726 & 70.6 & 55.4 & 66.8 \\
         & 18595 & 72.5 & 60.4 & 68.3 \\
         & 24912 & 72.9 & 51.5 & 68.3 \\
         & 4002 & 73.3 & 56.3 & 68.8 \\
         & 13263 & 70.0 & 50.8 & 67.7 \\
         & 23242 & 75.0 & 50.4 & 68.2 \\
         & 13839 & 72.5 & 56.7 & 68.9 \\
         & 30367 & 71.7 & 60.8 & 67.2 \\
         & 2904 & 67.5 & 53.7 & 67.2 \\
         & 7716 & 74.2 & 56.7 & 65.8 \\
         & 8832 & 68.3 & 52.5 & 66.0 \\
         & 20000 & 71.7 & 53.3 & 67.6 \\
         & 15724 & 71.7 & 56.5 & 66.7 \\
        \bottomrule
    \end{tabular}
    \caption{Detailed experimental results corresponding to Figure~\ref{fig: figure3} (Section 2.3), showing model performance under fixed seed.}
    \label{tab: table10}
\end{table}

\subsection{Detailed Experimental Results for All Models in Section 2.8}
\label{appendix:A.5}

\begin{table}[H]
    \centering
    \small
    \begin{tabular}{lcccc}
        \toprule
        \multirow{2}{*}{Model} & Evaluation Dataset & Instruction Position & Tensor Parallelism &  Baseline \\
         & Fluctuation & Fluctuation & Fluctuation & Fluctuation \\
        \midrule
        DeepSeek-R1-Distill-Qwen-32B & 2.4 & 0.2 & - & 0.0 \\
        QwQ-32B & 0.9 & 0.7 & - & 0.0 \\
        Skywork-OR1-32B-Preview & 1.2 & 0.3 & - & 0.0 \\
        TinyR1-32B-Preview & 0.5 & 0.7 & - & 0.0 \\
        DeepSeek-R1-Distill-Qwen-14B & 2.4 & 0.3 & 0.8 & 0.1 \\
        DeepCoder-14B-Preview & 1.9 & 0.9 & 0.7 & 0.0 \\
        Light-R1-14B-DS & 1.4 & 0.4 & 1.0 & 0.0 \\
        DeepSeek-R1-Distill-Qwen-7B & 0.8 & 0.0 & 1.3 & 1.4 \\
        Light-R1-7B-DS & 1.2 & 0.9 & 0.7 & 0.0 \\
        Skywork-OR1-Math-7B & 1.4 & 2.4 & 1.0 & 0.0 \\
        DeepSeek-R1-Distill-Qwen-1.5B & 2.0 & 2.0 & 1.5 & 0.0 \\
        DeepScaleR-1.5B-Preview & 2.1 & 2.2 & 1.4 & 0.0 \\
        Open-RS1 & 2.2 & 1.3 & 1.9 & 0.0 \\
        Open-RS2 & 0.3 & 1.4 & 1.3 & 0.0 \\
        Open-RS3 & 1.2 & 0.0 & 0.0 & 0.0 \\
        DeepCoder-1.5B-Preview & 1.5 & 0.2 & 0.4 & 0.0 \\
        ZR1-1.5B & 2.0 & 3.0 & 1.1 & 0.0 \\
        OpenRS-GRPO & 1.3 & 0.6 & 0.8 & 0.0 \\
        FastCuRL-1.5B-Preview & 0.7 & 0.4 & 1.0 & 0.0 \\
        STILL-3-1.5B-preview & 1.3 & 2.0 & 0.9 & 0.0 \\
        \bottomrule
    \end{tabular}
    \caption{AIME24}
    \label{tab: table11}
\end{table}

\begin{table}[H]
    \centering
    \small
    \begin{tabular}{lcccc}
        \toprule
        \multirow{2}{*}{Model} & Evaluation Dataset & Instruction Position & Tensor Parallelism &  Baseline \\
         & Fluctuation & Fluctuation & Fluctuation & Fluctuation \\
        \midrule
        DeepSeek-R1-Distill-Qwen-32B & 3.9 & 1.8 & - & 0.0 \\
        QwQ-32B & 2.7 & 0.4 & - & 0.0 \\
        Skywork-OR1-32B-Preview & 1.4 & 0.2 & - & 0.0 \\
        TinyR1-32B-Preview & 1.0 & 0.1 & - & 0.1 \\
        DeepSeek-R1-Distill-Qwen-14B & 3.3 & 0.9 & 0.2 & 0.1 \\
        DeepCoder-14B-Preview & 1.2 & 0.9 & 1.0 & 0.8 \\
        Light-R1-14B-DS & 2.5 & 0.1 & 0.1 & 0.4 \\
        DeepSeek-R1-Distill-Qwen-7B & 1.5 & 1.3 & 0.2 & 0.4 \\
        Light-R1-7B-DS & 3.8 & 2.0 & 2.2 & 0.0 \\
        Skywork-OR1-Math-7B & 2.5 & 1.0 & 0.5 & 0.0 \\
        DeepSeek-R1-Distill-Qwen-1.5B & 1.2 & 0.2 & 0.3 & 0.0 \\
        DeepScaleR-1.5B-Preview & 0.8 & 0.3 & 1.2 & 0.0 \\
        Open-RS1 & 1.3 & 0.4 & 0.2 & 0.0 \\
        Open-RS2 & 1.8 & 0.1 & 0.2 & 0.0 \\
        Open-RS3 & 1.6 & 0.9 & 1.4 & 0.0 \\
        DeepCoder-1.5B-Preview & 0.8 & 0.6 & 0.5 & 0.0 \\
        ZR1-1.5B & 0.8 & 1.2 & 1.2 & 0.0 \\
        OpenRS-GRPO & 2.8 & 0.5 & 1.1 & 0.0 \\
        FastCuRL-1.5B-Preview & 1.1 & 0.4 & 0.6 & 0.0 \\
        STILL-3-1.5B-preview & 1.7 & 0.4 & 0.4 & 0.0 \\
        \bottomrule
    \end{tabular}
    \caption{AIME25}
    \label{tab: table12}
\end{table}

\begin{table}[H]
    \centering
    \small
    \begin{tabular}{lcccc}
        \toprule
        \multirow{2}{*}{Model} & Option Bias & Correct Answer Bias & Tensor Parallelism &  Baseline \\
         & Fluctuation & Fluctuation & Fluctuation & Fluctuation \\
        \midrule
        DeepSeek-R1-Distill-Qwen-32B & 7.3 & 6.1 & - & 0.1 \\
        QwQ-32B & 11.9 & 3.4 & - & 0.1 \\
        Skywork-OR1-32B-Preview & 9.8 & 5.3 & - & 0.1 \\
        TinyR1-32B-Preview & 9.3 & 4.1 & - & 0.1 \\
        DeepSeek-R1-Distill-Qwen-14B & 9.4 & 3.2 & 0.6 & 0.7 \\
        DeepCoder-14B-Preview & 9.5 & 1.1 & 0.3 & 0.1 \\
        Light-R1-14B-DS & 8.4 & 2.4 & 0.2 & 0.0 \\
        DeepSeek-R1-Distill-Qwen-7B & 12.3 & 5.5 & 0.4 & 0.3 \\
        Light-R1-7B-DS & 10.8 & 2.6 & 0.4 & 0.4 \\
        Skywork-OR1-Math-7B & 14.4 & 6.3 & 0.1 & 0.4 \\
        DeepSeek-R1-Distill-Qwen-1.5B & 9.5 & 8.6 & 0.5 & 0.1 \\
        DeepScaleR-1.5B-Preview & 11.9 & 10.0 & 0.7 & 0.0 \\
        Open-RS1 & 7.7 & 7.1 & 0.2 & 0.0 \\
        Open-RS2 & 9.1 & 8.0 & 0.3 & 0.3 \\
        Open-RS3 & 9.2 & 8.9 & 0.0 & 0.0 \\
        DeepCoder-1.5B-Preview & 16.0 & 12.4 & 1.3 & 0.0 \\
        ZR1-1.5B & 8.9 & 10.9 & 0.3 & 0.0 \\
        OpenRS-GRPO & 5.0 & 2.6 & 0.0 & 0.0 \\
        FastCuRL-1.5B-Preview & 9.2 & 10.9 & 0.6 & 0.0 \\
        STILL-3-1.5B-preview & 5.5 & 3.6 & 1.2 & 0.1 \\
        \bottomrule
    \end{tabular}
    \caption{GPQA Diamond}
    \label{tab: table13}
\end{table}
\end{document}